\pdfoutput=1

\documentclass[11pt]{article}

\usepackage{acl}

\usepackage{times}
\usepackage{latexsym}
\usepackage{multirow}
\usepackage[T1]{fontenc}

\usepackage[utf8]{inputenc}
\usepackage{tcolorbox}
\usepackage[titletoc,title]{appendix}
\usepackage{colortbl}
\usepackage{microtype}
\usepackage[utf8]{inputenc} 
\usepackage[T1]{fontenc}    
\usepackage{amsfonts}       
\usepackage{amsmath}
\usepackage{amssymb}
\usepackage{mathtools}
\usepackage{amsthm}
\usepackage{bm}
\usepackage{color}
\usepackage{soul}           
\usepackage{url}
\usepackage{booktabs}       
\usepackage{nicefrac}       
\usepackage{enumerate}
\usepackage{graphicx}
\DeclareGraphicsExtensions{.pdf,.png,.jpg}
\graphicspath{{figures/}}
\usepackage{natbib}
\usepackage{comment}
\usepackage{hyperref}
\usepackage{centernot}     
\usepackage{algorithm}
\usepackage{algpseudocode}

\newcommand{\Exp}[2][]{\ensuremath{\mathbb{E}_{#1}\left[#2\right]}}                
\newcommand{\parg}{\makebox[1ex]{$\mathbf{\cdot}$}}                              
\newcommand{\mquad}{\kern-1em}													

\newcommand{\InNorm}[1]{{\left\vert\kern-0.2ex\left\vert\kern-0.2ex\left\vert #1 
    \right\vert\kern-0.2ex\right\vert\kern-0.2ex\right\vert}}                    

\newcommand{\InNormII}[1]{{\left\vert\kern-0.2ex\left\vert\kern-0.2ex\left\vert #1 
    \right\vert\kern-0.2ex\right\vert\kern-0.2ex\right\vert}_2}                    

\newcommand{\InNormInfty}[1]{{\left\vert\kern-0.2ex\left\vert\kern-0.2ex\left\vert #1 
    \right\vert\kern-0.2ex\right\vert\kern-0.2ex\right\vert}_{\infty}}           

\newcommand{\Abs}[1]{\ensuremath{\left \lvert #1 \right \rvert}}                 
\DeclarePairedDelimiterX{\Inner}[2]{\langle}{\rangle}{#1, #2}                    

\newcommand{\MI}{\mathnormal{I}}                                                     

\newcommand{\Set}[1]{\{#1\}}                                                     

\definecolor{gray}{rgb}{0.7,0.7,0.7}

\newcommand{\TODO}[1]{{\color{red}TODO:}\emph{#1}}

\newtheorem{theorem}{Theorem}

\theoremstyle{definition}
\newtheorem{example}{Example}[section]

\input{tikz}
\input{tikz}
\usetikzlibrary{graphs}
\usetikzlibrary{quotes}

\tikzset{
nodestyle/.style={
  ellipse,
  inner sep=0pt,
  minimum size=8mm,
  align=center,
  font=\bf,
  draw=black,
  fill=white
  }
}

\tikzset{
nodestyle1/.style={
  circle,
  inner sep=0pt,
  align=center,
  text=white,
  font=\bf,
  draw=white,
  fill=black!80
  }
}

\newcommand{\answersumm}{\textsc{AnswerSumm}}
\newcommand{\td}{\textsuperscript{\textdagger}}
\newcommand{\bartscore}{\textsc{BARTScore}}

\newcommand{\grayrule}{\arrayrulecolor{gray}\noalign{\vskip 1mm} \hline \noalign{\vskip 1mm}}

\newcommand{\dP}{\mathbb{P}}

\title{Improving Faithfulness of Abstractive Summarization by
Controlling Confounding Effect of Irrelevant Sentences}

\author{Asish Ghoshal, Arash Einolghozati, Ankit Arun, \\
    {\bf Haoran Li}, {\bf Lili Yu}, {\bf Vera Gor}, {\bf Yashar Mehdad}, \\
    {\bf Scott Wen-tau Yih} \and {\bf Asli Celikyilmaz} \\
    Meta AI, Seattle, Washington, USA
}

\begin{document}
\maketitle
\begin{abstract}
Lack of factual correctness is an issue that still plagues state-of-the-art summarization systems despite their impressive progress on generating seemingly fluent summaries. 
In this paper, we show that factual inconsistency can be caused by irrelevant parts of the input text, which act as confounders.  
To that end, we leverage information-theoretic measures of causal effects to quantify the amount of confounding and precisely quantify how they affect the summarization performance. 
Based on insights derived from our theoretical results, we design a simple multi-task model to control such confounding by leveraging human-annotated relevant sentences when available. Crucially, give a principled characterization of data distributions where such confounding can be large thereby necessitating obtaininig and using human annotated relevant sentences to generate factual summaries.
Our approach improves faithfulness scores by 20\% over strong baselines on \answersumm{} \citep{fabbri2021answersumm}, a conversation summarization dataset where lack of faithfulness is a significant issue due to the subjective nature of the task. Our best method achieves the highest faithfulness score while also achieving state-of-the-art results on standard metrics like ROUGE and METEOR. We corroborate these improvements through human evaluation.
\end{abstract}

\begin{figure}[t]
    \centering
    \begin{tcolorbox}[boxsep=0pt,left=0pt,colframe=white]
    \scriptsize
    \begin{tabular}{p{\linewidth}}
    \textbf{Question}: Every fall when the weather gets cooler it is a given that a mouse will end up in my house. I understand that this happens probably to most home owners ... But if I wanted to take some action next summer/early fall to prevent mice from entering, what could I do?\\
    \grayrule
    \textbf{Answers}:\\
    1. \textbf{This is probably a futile exercise}. Mice can squeeze through the smallest of gaps so you'd have to virtually hermetically seal your house to prevent any mouse coming in... \textbf{What you can do is make your house a less welcoming environment}. ...\\
    2. When I bought my house it had a big problem with mice. … What I did was: \textbf{make sure there was no accessible food} I checked every place in the house that a mice could hide behind. … 
    6. \textbf{I've found no solutions to keeping them out, nor will I use poison, as I've no desire to see my dog or our neighbor's cat poisoned too. You can try to have your house sealed, but unless the seal is virtually hermetic, they will find their way into the warmth of your home}. And mice can chew holes in things, so maintaining the perfect seal will be tough. ...\\
    7. ... get some kitty pee : If you prefer, mice are supposed to hate the smell of peppermint and spearmint, so there are repellents made from mint :\\
    8. I was extremely skeptical of electronic pest repellers and I have no idea whether they work with rodents. [\dots] work very well against roaches. ...  \\
    \grayrule
    \textbf{Human Summary}: Whilst you might attempt to seal your house as best as possible, the general consensus is that this is an almost impossible task. However, it is possible to make your home less attractive to mice by making sure that food is properly stored and out of reach. Mouse poison could be a potential solution, but there is the risk of poisoning other animals. A rather unusual solution could be to attract birds of prey to your garden, although, it's unclear how you could do this.\\
    \grayrule
    \textbf{BART}: The most effective way to stop mice from gaining access to your house is to block up any potential access holes or cover them with plastic. Failing this, you could also try putting out poison.
    \end{tabular}
    \end{tcolorbox} 
    \vspace{-0.5cm}
    \caption{\small An example instance from the \answersumm{} dataset 
    \cite{fabbri2021answersumm} derived from Stack Exchange. For a given a question and a list of community answers, a human written summary and relevant sentence annotations (in bold) are shown. The BART \cite{lewis2020bart} generated summary combines information from two different sentences ``keep[ing] all food sealed in plastic containers'' and ``Sealing basement window cracks with silicone'' to generate the summary ``cover holes with plastic,'' which is not faithful since the answers suggest covering/sealing holes with silicone and not plastic.}  
    \label{fig:example}
    \vspace{-0.5cm}
\end{figure}
\section{Introduction}
Large neural language models like BART \cite{lewis2020bart} and T5 \cite{raffel2019exploring} have spurred rapid progress on the problem of abstractive summarization.
While such models can generate fluent summaries, their practical deployment is hampered by lack of confidence in the factual correctness of generated summaries, where models can \emph{hallucinate} facts that are not supported by the input document. Generating factual summaries is especially challenging in conversational domains like chat and social media \cite{tang2021confit}. The problem is further exacerbated by the lack of reliable metrics for evaluating factuality or faithfulness of generated summaries. Commonly used metrics like ROUGE are known to correlate poorly with human judgements of factuality \cite{gabriel-etal-2021-go, falke2019ranking}.

Recent work  \cite{xie2021factual} on evaluating faithfulness of abstractive summarization systems posit that the hallucination in abstractive summarization systems is caused by the language prior, where models rely on prior information encoded in the language model to generate parts of the summary, as opposed to the information contained within the document. Towards that end, \citet{xie2021factual} measure (un-)faithfulness by quantifying the causal effect of the language prior on the generated summary. However, when a document contains many irrelevant or distractor sentences, as is typically the case in many conversation summarization datasets like QMSum \cite{zhong2021qmsum} and AnswerSumm \cite{fabbri2021answersumm}, factual mistakes can still occur even when the model relies on the information contained in the input document. We focus on \emph{community question answering (CQA) summarization} task where the goal is to summarize the answers for a given question which are often opinionated. One particular dataset we use to validate our approach is AnswerSumm \cite{fabbri2021answersumm} which contains human annotations of relevant sentences thereby making it especially suitable to understand how irrelevant sentences can act as confounders and cause hallucinations in abstractive summarization systems. Figure \ref{fig:example} shows an example from the dataset along with the summary generated by BART \cite{lewis2020bart} which suffers from intrinsic hallucination. 

\paragraph{Contribution.}
We rigorously study the effect of irrelevant sentences in a document on the quality of the generated summary.
A key contribution of our work is using recently developed information-theoretic measures of causality \cite{ay2008information} to quantify the confounding effect of irrelevant sentences. We show that confounding reduces the log-likelihood of generated summaries thereby affecting faithfulness\footnote{The relationship between generation probability and faithfulness has been previously explored by \citet{yuan2021bartscore} in developing the \bartscore{} metric.}. We also characterize summarization tasks where the confounding can be small and therefore ignored, or large thereby necessitating methods requiring confounding control to improve faithfulness. 

Towards that end, we design a simple extract-and-generate summarization model where extraction of relevant sentences is decoupled from generation of the abstractive summaries. The first step (relevant sentence extraction) does not introduce factual errors and by training the abstractive generation stage on human annotated relevant sentences we reduce confounding due to irrelevant sentences. Our method achieves the best test log-likelihood scores among over strong baselines on the AnswerSumm dataset while matching BART on ROUGE scores. It also improves the faithfulness score by 20\% as evaluated by an adapted \bartscore{}. We verify that the improvement in the log-likelihood (and consequently faithfulness) is due to the confounding control by estimating the amount of confounding in each example. For the top-50 examples in the test set with highest confounding, we find statistically significant improvement in log-likelihood scores of our model over the vanilla BART model as compared to the bottom-50 examples with the least amount of confounding, whereas there is no difference between ROUGE scores between the two sets.
We corroborate our results through human evaluation and show that our proposed method matched or improved faithfulness over the BART baseline on 72\% of examples. 

Our work contributes towards theoretical understanding of abstractive summarization systems by identifying, quantifying, and measuring additional sources of hallucination in addition to the language prior identified in \cite{xie2021factual}, as well as developing simple yet effective solutions towards mitigating such effects. 
\section{Causal Analysis}
\label{sec:causal_analysis}

\begin{figure}
    \centering
\begin{tikzpicture}[every node/.style={scale=0.9}]
\node[nodestyle] (nX)  {$\Set{Q, X}$};
\node[nodestyle, right=1cm of nX] (nXR) {$\Set{Q, X_R}$};
\node[nodestyle, right=1cm of nXR] (nY) {$Y$};
\draw (nX) edge[->] (nXR);
\draw (nXR) edge[->] (nY);
\end{tikzpicture}%
    \caption{A causal generative model for extractive-abstractive summarization.}
    \label{fig:causal_model}
\end{figure}


In this work, we propose an extractive-abstractive summarization model and argue that it is a more \emph{faithful} summarization approach, compared to a direct model that implicitly performs extraction and abstractive generation.
This section provides the theoretical justification by formulating a causal generative model for extractive-abstractive summarization.


\subsection{Setup}

Let $X$ be an input document, $Q$ be a query and $X_R \subseteq X$ denote sentences in $X$ that are relevant to the query $Q$. Figure~\ref{fig:causal_model}
shows a simplified causal generative model for EA summarization. Given a query and the document, an extractor function $e$ selects relevant sentences and then another function $g$  generates the final summary from the query and the relevant sentences. The generative process is formalized by the following structural equation model~(SEM)~\cite{pearl2012causal},
where $\epsilon$'s are independent noise or exogenous variables:
\begin{align}
    X_R &= e(Q, X, \varepsilon_1) \notag \\
    Y &= g(Q, X_R, \varepsilon_2) \label{eq:sems}
\end{align}
We do not model the effect of the language model prior on the final summary and include it in the exogenous variable affecting the summary ($\varepsilon_2$).
The true function to generate the final summary from the input is given by the composite function $f = g  \circ e$. Note that all three functions $f$, $g$ and $e$ can be modeled as seq2seq tasks, and Transformer models are expressive enough to approximate all these functions \cite{yun2019transformers}. Therefore, it might appear that there is no difference between learning the function $f$ directly from data over learning $g$ and $e$ separately and then composing them. In the following sections, we show that this
is not the case.

\subsection{Causal effect of irrelevant sentences}
First, we quantify the causal effect of the irrelevant sentences $CE(X_{R^c})$ on the summary $Y$ where $R^c$ denotes the complement of $R$. The information flow~\cite{ay2008information} from $X$ to $Y$ is defined as:
\begin{align*}
\MI(X \rightarrow Y) &= \sum_{x} p(x) \sum_{y} p(y \mid do(X=x)) \times \\
    &\qquad \log \frac{p(y \mid do(X=x))}{\sum_{x'} p(y \mid do(X=x'))}
\end{align*}
The above differs from the standard mutual information in that the observational distribution $p(y \mid x)$ is replaced by the interventional distribution $p(y \mid do(X=x))$. This ensures that the effect of $X$ on $Y$ is accurately measured after removing the effect of confounders on $X$ and $Y$.

Following \citet{xie2021factual} who remove the causal effect of language prior from the total causal effect, we quantify $CE(X_{R^c})$ as the difference between the total causal effect of $\Set{Q, X}$ on $Y$ and the direct causal effect of $\Set{Q, X_R}$ on $Y$:
\begin{align}
    &CE(X_{R^c}) = \notag \\
    &\quad \Abs{\MI(\Set{Q, X} \rightarrow Y) - \MI(\Set{Q, X_R} \rightarrow Y)}.
\end{align}

Our first theoretical result is the following theorem that shows that causal effect of the irrelevant sentences can be estimated from the observed data, i.e., without performing any interventions.
\begin{theorem}
\label{thm:causal_effect}
The causal effect of the irrelevant sentences on the final summary is $CE(X_{R^c}) = H(X_R \mid X, Q) - H(X_R \mid X, Q, Y) \geq 0$.
\end{theorem}
Full proofs of the theorems can be found in Appendix \ref{app:theoretical_result}.
Intuitively, the above result implies that if observing the summary makes identifying the relevant sentences significantly easy (equivalently reduces entropy) over identifying the relevant sentences from the input document alone then the \textit{irrelevant sentences} have large confounding effect on the summary. To better understand the implications of the above result, we provide some example summarization tasks with varying causal effect of the irrelevant sentences on the final summary.

\begin{example} 
\label{ex1}
Imagine all sentences were relevant, i.e., $X_R=X$, then $H(X_R \mid X, Q) = H(X_R \mid X, Q, Y) = 0$ and 
$CE(X_{R^c}) = 0$.
\end{example}

\begin{example} 
\label{ex2}
Imagine only the first sentence is relevant. Since there is a deterministic relationship between $X_R$ and $X$, i.e. $X_R = X_1$, once again, $H(X_R \mid X, Q) = H(X_R \mid X, Q, Y) = 0$ and  $CE(X_{R^c}) = 0$.
\end{example}

\begin{example} 
\label{ex3}
Imagine the relevant sentence is randomly assigned, i.e., $X_i=X_R$ where $1 \leq i \leq \Abs{X}$ is picked uniformly at random. Moreover, imagine that the final summary $Y$ is simply $X_R$. Then $H(X_R \mid X, Q) = \log \Abs{X}$ and $H(X_R \mid X, Q, Y) = 0$ and $CE(X_{R^c}) = \log \Abs{X}$.
\end{example}
The first two examples represent easy cases where the irrelevant sentences do not have any effect on the final summary and direct summarization from input document can work well. The third example represents the worst case where the causal effect of irrelevant sentences is the largest and can significantly affect the performance of a summarization model that generates the summary directly from the input document. Next, we give a precise characterization of how irrelevant sentences affect abstractive summarization performance.

\subsection{Effect of irrelevant sentences on abstractive summarization performance}
Here, we  show how $CE(X_{R^c})$ affects the generation probability (or equivalently the negative log-likelihood loss) of abstractive summarization. 
Let $\dP$ be the true data distribution over triples $(Q, X, R, Y)$ with probability mass function $p$ as determined by the structural equation model in~(\ref{eq:sems}). 
Also define the loss of predictors to be the expected negative log-likelihood of the predictions under the true data distribution:
\begin{align}
    l(f) &= - \Exp[(q,x,r,y) \sim \dP]{\log p(f(q, x) \mid q, x)} \\
    l(g) &= - \Exp[(q,x,r,y) \sim \dP]{\log p(g(q, x_r) \mid q, x_r)}
\end{align}
where $p(y \mid x)$ and $p(y \mid r)$ are the conditional distributions of $y$ under the true data distribution $\dP$. 

Our next result shows that expected loss of $f$ is always worse than that of $g$ and the difference between the two losses is precisely the causal effect of the irrelevant sentences $CE(X_{R^{c}})$:
\begin{theorem}
\label{thm:log_likelihood}
$l(f) = l(g) + CE(X_{R^c})$
\end{theorem}
This difference in likelihood is a function of the causal effect of the irrelevant sentences and can be small or large depending on the data distribution (Examples~\ref{ex1}-\ref{ex3}).
This does not necessarily hold for other metrics like ROUGE as we will  demonstrate empirically. The causal analysis presented here forms the basis for our approach, which we present in the next section. 
\begin{figure*}
    \centering
    \includegraphics[width=0.75\textwidth]{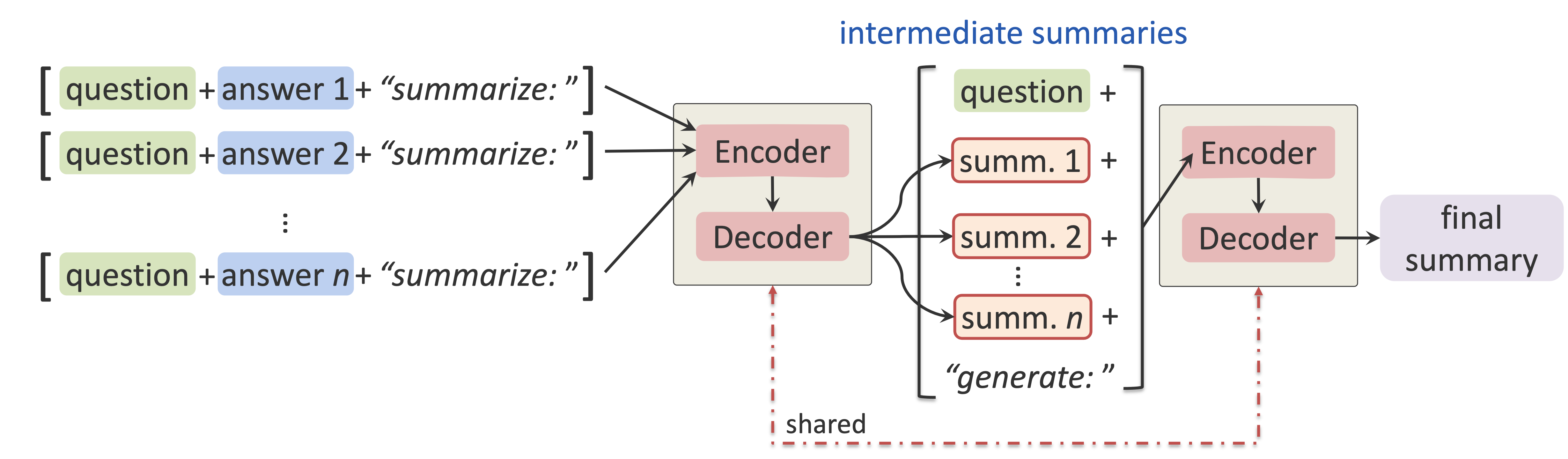}
    \caption{\small Repeated summarization model architecture showing the underlying seq2seq model unrolled over time. First each answer is given as input to the model along with the question and prefix ``summarize:'' to generate answer summary. After all answer summaries have been generated, they are passed through the seq2seq model again, this time with prefix ``generate:'' to generate the final summary.}
    \label{fig:architecture}
\end{figure*}
\section{Approach}
The previous theoretical results justify extract-and-generate approaches over direct summarization models when supervision for relevant sentences is available. This is especially important for opinionated conversation summarization which is closer to the setting in Example \ref{ex3} due to the inherent subjectivity in identifying relevant sentences \footnote{This is evidenced by low inter-annotator agreement (Fleiss Kappa of 0.25) for relevant sentence identification \cite{fabbri2021answersumm} in \answersumm{}.}. 
\textbf{Ensuring conditional independence of abstractive generation from the input passage given gold relevant sentences is crucial}. Using relevant sentences as additional guidance signal in a direct model, which is the approach followed in GSum \cite{dou2020gsum}, can still introduce confounding due to distractor sentences. Lastly, we also need to select the most important opinions for summarization and not simply summarize everything. As such, we design a seq2seq model that first extracts relevant sentences from each answer and then summarizes them. 

\subsection{SURE-Summ: Supervised Repeated Summarization for CQA summarization} 
Shifting our focus to community question answering summarization, we denote the set of answers for a question $Q$ by $X_1, \ldots, X_n$. Let $R_i$ denote the relevant sentences in the $i$-th answer. We use a single model for both the probability of the final summary given relevant sentences $p(Y|Q, R_1, \ldots, R_n)$ and the probability of a sentence being relevant given the answer $p(R_i|Q, X_i)$. Given recent success of casting different tasks into the same format for encouraging maximal sharing of information in a multi-task setup \cite{mccann2018natural, raffel2019exploring}, we cast both tasks in the same setup with different prompts or prefixes. We concatenate the question $Q$ and an $X_i$ together with the prefix ``\texttt{summarize:}'' to generate the relevant sentences for the i-th answer by casting it as an \emph{extractive summarization} task. We do not constrain the generation in anyway and by virtue of the training process, the generated summaries are almost extractive. To make the generated answer summaries fully extractive, we perform some minor post-processing which is described in the next paragraph. We use the prefix ``\texttt{generate:}'' to generate the final abstractive summary from the question and the concatenation of post-processed generated extractive summaries of all the answers. This process of repeatedly performing extractive summarization over answers to extract relevant sentences and a final abstractive summarization step to generate the final summary motivates the name of our method.

\noindent{\textbf{Extractive post-processing.}} Let $\widetilde{Y}_i$ denote the extractive summary generated by the model for the $i$-th answer $X_i$ and let $Y_i$ denote the output of the post-processing step. For each sentence $X_{i,j} \in X_i$ if $\mathrm{ROUGE1}_{\mathrm{precision}}(\widetilde{Y}_i, X_{i,j}) \geq 0.8$ then we include the sentence $X_{i,j}$ in $Y_i$.

\noindent{\textbf{Inference.}} During inference, we simply generate relevant sentences for each answer separately, apply the extractive post-processing step, and generate the final summary from the concatenation of the question and extractive answer summaries. 

\noindent{\textbf{Training.}} 
Given the gold relevant sentences $R_{i}$ for the $i$-th answer, and the reference summary $Y$, we train our seq2seq model by minimizing the following multi-task objective: 
\begin{align}
   \textstyle H(Y, P(Q, R)) + \sum_{i=1}^n H(R_i, P(Q, X_i)), \label{eq:loss}
\end{align}
where $P(Q, X) = p(\parg \mid Q, X)$ denotes the model predicted probabilities of the output sequence, and $H(\parg, \parg)$ denotes the cross-entropy loss. Furthermore, to reduce the memory consumption during the training, we sample a single answer,
with probability proportional to the fraction of relevant sentences in the answer, to 
approximate the summation in the second term in \eqref{eq:loss}. 

\subsection{Pipeline Models}
We also experiment with pipeline models by training separate extractor and generator models.  The generator is trained separately to generate the final summary from the gold relevant sentences while the extractor is trained to predict the relevant sentences given the answers. During inference, the generator is provided with \emph{predicted relevant sentences} produced by the extractor as input. 
We use the SURE-Summ model trained as before as a powerful relevant sentence extractor while a standard pre-trained transformer model (BART-large) as the generator.
While this approach ensures conditional independence of the generated summary from the original answers, it also suffers from the exposure bias due to the use of predicted relevant sentences at inference time. 

\section{Metrics}
\subsection{Automatic Metrics}
To evaluate our models, we report ROUGE~\cite{lin-2004-rouge} and METEOR \cite{banerjee2005meteor} scores which have been widely used in the summarization literature. We also report test log-likelihood  
and BARTScore~\cite{yuan2021bartscore} (as modified for our setting) for evaluating \textit{faithfulness}. 
Our analysis on using BERTScore~\cite{BERTScore} yielded saturated scores for our task where all models scored in excess of 95 F1-score.

\subsection{Adapting \textbf{\bartscore{} for EA summarization.}}
\label{anewfaithfullnessmetric}
There are two problems with directly using the vanilla pre-trained \bartscore{} provided by \citet{yuan2021bartscore}. First, the vanilla \bartscore{} is trained on news articles (i.e., CNN/DM datasets) while we focus on CQA data, which has a different discourse structure and semantic features compared with the news datasets. As suggested by \citet{yuan2021bartscore}, the domain shift can be mitigated by fine-tuning the underlying BART model on the task data. Second, even if we were to train the underlying BART model in \bartscore{} on the task data, as described in Section~\ref{sec:causal_analysis}, its scores (log-likelihood) would be affected by confounding due to the irrelevant sentences. Here, we describe a principled approach to evaluate faithfulness using \bartscore{} when gold relevant sentence annotations are available.

We first finetune BART to generate the final summary from golden relevant sentences. This oracle model is not affected by confounding due to the irrelevant sentences and provides more reliable log-likelihood scores. For extract-and-generate models, we measure the faithfulness as the log-likelihood score of the generated summary computed by the aforementioned BART model using the \emph{predicted relevant sentences} as the input. For direct models, we evaluate faithfulness of the generated summary against the entire input document. In essence,
the faithfulness metric measures the faithfulness of the final summary with respect to the sentences that were actually used in generating the summary. By doing so, we separate estimation of the factuality of the final (abstractive) summary from the quality of relevant sentences. Note that a faithful summary does not indicate a relevant or useful summary. 

\subsection{Human Evaluation}
We conduct human evaluations and use multiple raters to evaluate summaries on a five point Likert scale along five axes: faithfulness, relevance, coherence, multiperspectivity, and overall quality (details in Appendix \ref{app:human_eval}).

\section{Experiments}
\paragraph{Dataset}
We use the \answersumm{} dataset \cite{fabbri2021answersumm} which contains human-annotated relevant sentences for each example. More details can be found in Appendix \ref{app:dataset_details}.

\paragraph{Baselines}
\label{sec:baselines}
We evaluate two types of models: \textit{direct models}, which ignore the relevant sentence annotations and directly generate the summary from the answers, and \textit{supervised extractive-abstractive models}, which use the relevant-sentence annotation and decouple the relevant sentence extraction from the abstractive generation.

The direct models include BART, T5 and PEGASUS~\cite{zhang2020pegasus}. 
We finetuned these pre-trained seq2seq models on the \answersumm{} dataset. We used BART-large and T5-base versions for all our experiments because we were unable to train T5-large even with batch size 1 due to the lack of sufficient GPU memory. 
For the supervised extractive-abstractive baseline, Supervised RoBERTa-BART, we train RoBERTa \cite{liu2019roberta} to extract relevant sentences and used a BART model trained on gold relevant sentences as the generator. 

\paragraph{Experimental setup}
All our experiments were performed on AWS P4d instances running Nvidia A100 GPUs with 40GB GPU memory. Models were trained for 50 epochs with an effective batch size of 32 (more details in Appendix \ref{app:experiment_details}).

\begin{figure*}
    \centering
    \includegraphics[width=0.9\linewidth]{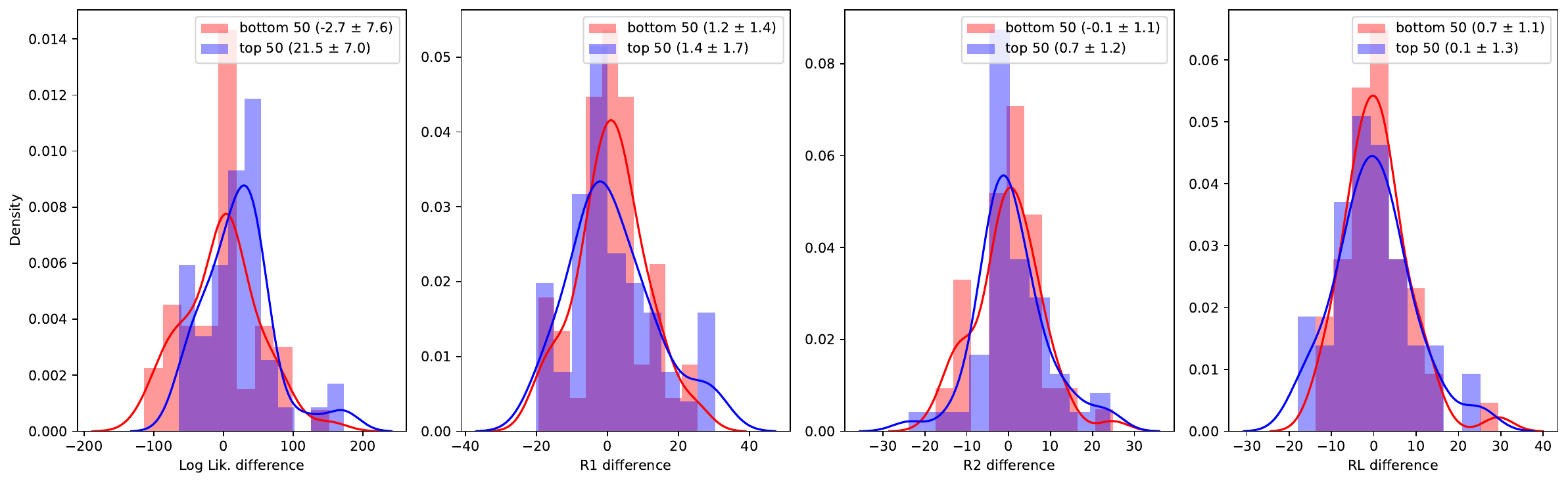}
    \caption{Difference between test set log-likelihood scores (scaled by 100) of SURE-Summ and BART for top 50 (respectively bottom 50)  examples with highest (respectively lowest) causal effect of irrelevant sentences ($CE(X_{R^c})$). Difference between ROUGE scores of the two methods for the same set of examples are shown for comparison. The mean and standard errors of the difference between scores in the two groups (top/bottom-50) are shown within braces.}
    \label{fig:answersumm_causal_effect}
\end{figure*}%
\begin{table*}[htbp]
    \centering
    \small
    \begin{tabular}{ccccccc}
        \hline
        Method & \#Params & F1 Rouge 1/2/L & Meteor & Faithfulness & Log Lik. & Length \\
        \hline
        \hline
        Oracle BART & 406M & 34.2 / 11.6 / 23.9 & 24.4 & -87.1 & -362.8 & 214 \\
        \hline
        \multicolumn{7}{c}{\textbf{Supervised extractive-abstractive models}} \\
        \hline
        SURE-Summ & 406M & 30.2 / 8.9 / 20.8 & 21.4 & \textbf{-125.6} & -364.9 & 215 \\        
        Sup. RoBERTa-BART & 761M & 30.6\td / 9\td / 20.9\td & 22\td & -86\td & -399.2 & 242 \\
        SURE-Summ-BART & 812M & {\bf 30.8} / {\bf 9.3} / \textbf{21.1} & {\bf 22.4} & {\bf -79.9} & -393.9 & 240 \\
        \hline
        \multicolumn{7}{c}{\textbf{Direct models}}\\
        \hline
        BART & 406M & 30.2\td / \textbf{9.2} / \textbf{21.3} & 20.3 & -156.9 & -372.1\td & 184 \\
        PEGASUS & 568M & 29.6 / 8.8 / 20.7  & 20.4 & -153.8 & -378.9 & 198 \\
        T5  & 220M & 29.5 / 8.8 / 20.6 & 20.6\td & -201.8 & -391.2 & 218 \\
        \hline
    \end{tabular}
    \vspace{-0.2cm}
    \caption{\small Performance of different methods on the test set of the \answersumm{} 
    data set. The Oracle model is trained on gold relevant sentences and also uses gold relevant sentences during inference to generate the final summary. }
    \label{tab:main_results}
\end{table*}
\begin{table}[]
    \centering
    \resizebox{\linewidth}{!}{%
    \begin{tabular}{c@{}c@{}c@{}c@{}}
    \hline
    & \textbf{SURE-Summ-BART } \ \ &\textbf{ RoBERTA-BART} \ \ &\textbf{SURE-Summ} \\
    & better / worse / same &  better / worse / same  &  better / worse / same  \\
    \hline
    Faithfulness \ \ & 52.6 / 26.3 / 21.1  & 41.0 / 59.0 / 00.0 & 53.2 / 27.7 / 19.1 \\
    Relevance & 31.6 / 31.6 / 36.8  & 23.1 / 48.7 / 28.2 & 19.1 / 44.7 /36.2 \\
    Coherence & 39.5 / 34.2 / 26.3  & 35.9 / 41.0 / 23.1 & 38.3 / 34.0 / 27.7 \\
    Overall & 42.1 / 36.8 / 21.1  & 51.3 / 38.5 / 10.3 & 51.1 /34.0 / 14.9\\
    \hline
    \end{tabular}}
    \vspace{-0.2cm}
    \caption{\small Human evaluation of the models against the BART baseline.}
    \label{tab:human_eval_results1}
\end{table}%
\section{Results}
\label{sec:results}

\subsection{Automatic evaluation}

Table \ref{tab:main_results} shows the performance of our model vis-\`{a}-vis other methods on the test set of the \answersumm{} dataset.
We see that all the extractive-abstractive models achieve significantly better faithfulness scores over direct methods as evaluated by our modified \bartscore{}. SURE-Summ improves faithfulness scores by 20\% over the vanilla BART model while having the same number of parameters. Furthermore, the length of summaries produced by SURE-Summ is almost the same as the Oracle BART model. 

\subsection{Human evaluation}
Table \ref{tab:human_eval_results1} shows the results from human evaluation for the two pipeline models and the SURE-Summ model. We compute the fraction of examples where the given model scored better, worse, or equal to the standard BART model along the four axes of evaluation. The SURE-Summ model received better or equal score than the BART baseline on 72\% of examples on faithfulness. Also, the overall  quality was better or equal to the BART baseline for 66\% of examples as evaluated by humans.

\begin{figure}
    \centering
    \includegraphics[width=0.8\linewidth]{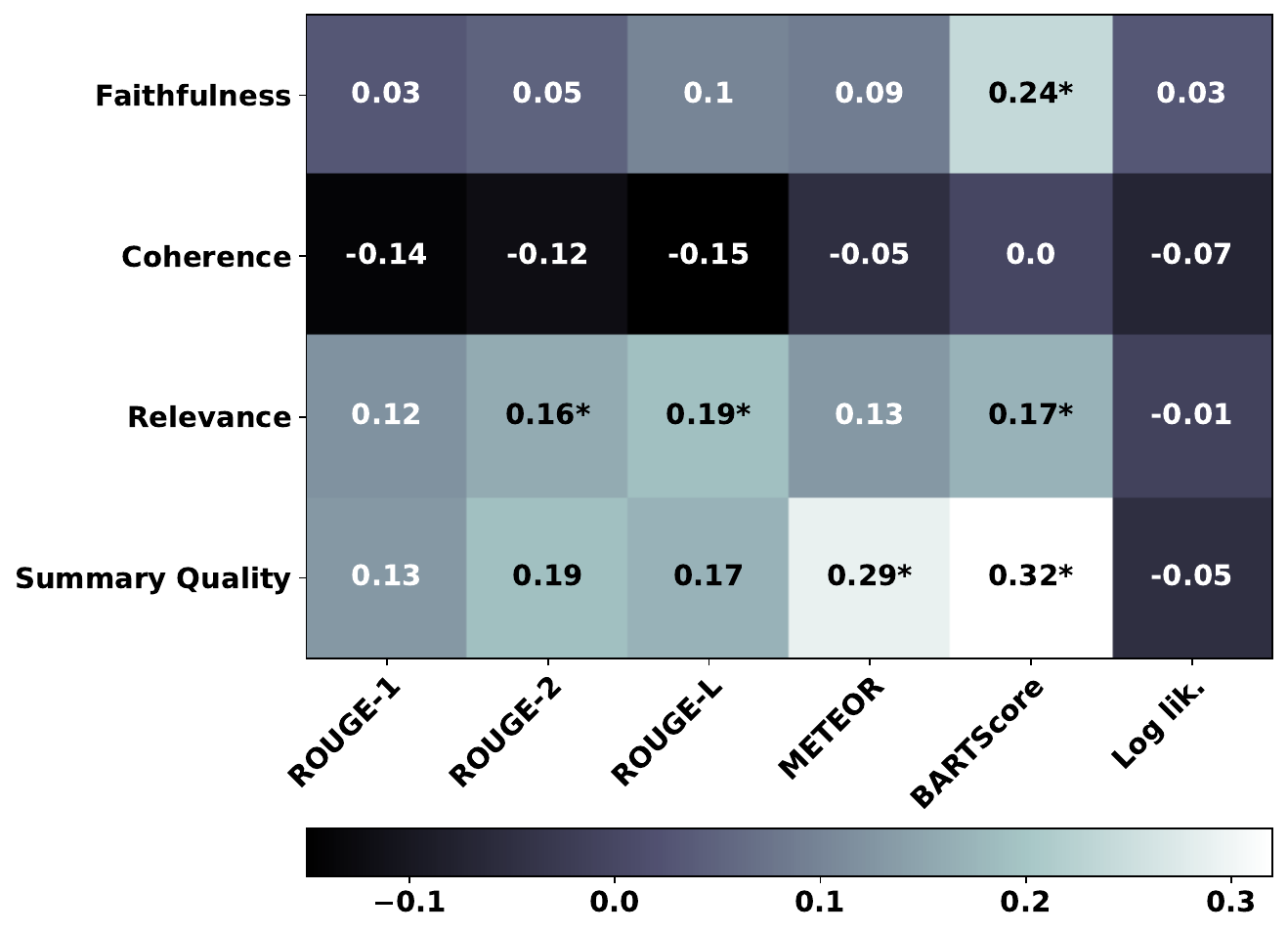}
    \caption{Spearman correlation of automatic metrics with human scores.}
    \label{fig:metrics_correlation}
\end{figure}

\paragraph{Correlation with automatic metrics}
Figure \ref{fig:metrics_correlation} shows the Spearman correlation between various automatic metrics used in the paper with human scores for four models: RoBERTa-BART, SURE-Summ-BART, Sup. Oracle BART, and BART. Asterisk denote statistically significant correlation at p-value of 0.05.
Among all metrics, \bartscore{} is the most positively correlated with all human metrics except for relevance and is the only metric that has statistically significant correlation with faithfulness. We note that our version of \bartscore{} formulation was not designed to measure the relevance. These results demonstrate that \bartscore{} is the best metric to evaluate faithfulness and coherence, while ROUGE is a better measure of relevance. Lastly, while it appears that in absolute terms our faithfulness metric is poorly correlated with human judgements, this in line with other metrics like FEQA \cite{durmus2020feqa} where the maximum correlation across two summarization datasets (CNN/DM and XSum) was 0.32. Similarly, the maximum correlation of the proposed metrics in \cite{peyrard2019simple} is less than 0.3.%

\subsection{Analysis}
\paragraph{Measuring the confouding in the \answersumm{} dataset.}
Following the Theorem \ref{thm:causal_effect}, we measure the amount of confounding due to the irrelevant sentences in the \answersumm{} dataset by training two binary classification models to estimate $H(X_R \mid X, Q)$ and $H(X_R \mid X, Q, Y)$ respectively. We train two RoBERTa models for this purpose. The RoBERTa model that predicts relevant sentences from $X$ achieves an F1 score of 53.9 while the model that uses the observed summary achieves an F1 score of 57.5 indicating significant confounding. We then estimate the causal effect of the irrelevant sentences for each example in the test set as per Theorem \ref{thm:causal_effect}.  Fig. \ref{fig:answersumm_causal_effect} shows the difference in log-likelihood scores\footnote{We multiply log-likelihood scores by 100 to have them in roughly the same range as other metrics.} computed by SURE-Summ and BART. We have shown the results for the top-50 and the bottom-50 examples base on the confounding. For control, we also compute the difference in ROUGE scores computed by the two methods for the same set of examples (top/bottom 50). We see that SURE-Summ improves the log-likehood (statistically significant at p-value of 0.05 with a two-sample T-statistic of 2) for the examples with highest amount of confounding, whereas there is no (statistically significant) difference between the ROUGE scores. 

\paragraph{Multi-task objective improves accuracy of relevant sentence prediction.}
Table \ref{tab:rel_sent_pred} shows the ROUGE F1 score achieved by the SURE-Summ model for predicting relevant sentences and a supervised RoBERTa model. We see that the information sharing due the multi-task objective results in an improvement of 6.9\% (R1 score) for relevant-sentence selection in the SURE-Summ model. In fact, the performance of SURE-Summ for predicting relevant sentences matches the RoBERTa model which uses the gold summary to predict relevant sentences during the inference as described in the previous paragraph. This further vindicates our unified seq2seq multi-task objective and.  

\paragraph{Using SURE-Summ in the pipeline model produces more relevant and faithful summaries.}
The previous result motivated us to use the SURE-Summ model purely as a relevant sentence predictor in a pipeline model. The relevant sentences predicted by the SURE-Summ model is then passed as input to a BART model, trained on gold relevant sentences, to generate an abstractive summary of the (predicted) relevant sentences. This model achieves the best of both worlds: generating summaries that has the highest faithfulness scores among all models while also producing the most relevant summaries as evidenced by the ROUGE scores which are also the best among models. Note that ROUGE scores primarily measure relevance since they compute token match between the predicted and gold summaries. This obviously comes at the cost of doubling the number of parameters. 
\begin{table}[]
    \centering
    \begin{tabular}{cc}
    \hline
    Method &  F1 Rouge 1/2/L \\
    \hline
    SURE-Summ & 57.6 / 50.2 / 52.6 \\
    RoBERTa & 53.9 / 47.4 / 49.3 \\
    \hline
    \end{tabular}
    \caption{\small Performance of models for predicting relevant sentences which we view as an extractive summarization task. }
    \label{tab:rel_sent_pred}
\end{table}

\section{Related Work}

\paragraph{Evaluating and improving faithfulness}
Various metrics have been proposed to evaluate the faithfulness of abstractive summarization models, such as QA-based methods \cite{durmus2020feqa} and entailment-based methods \cite{maynez2020faithfulness}. \citet{durmus2020feqa} posit that there is an abstractiveness-faithfulness trade-off, where more abstractive summaries are typically less faithful.
Unfortunately, automatic evaluation of the faithfulness is still an open problem \citep{surveyevaltextgen}. For instance,~\citet{falke2019ranking} and \citet{gabriel-etal-2021-go} show that out-of-the-box entailment models trained on NLI datasets fail to evaluate the summarization factuality. The most similar to our work is of \citet{xie2021factual}, who propose measuring the faithfulness by estimating the causal effect of the language prior on the final summary. However, the goal in this paper is to not merely measure but also improve the faithfulness.
More recently, \citet{yuan2021bartscore} showed that the log-likelihood scores of strong generative models like BART (referred to as \bartscore{} in their paper) can correlate well with human generated factuality scores. Motivated by this, we use \bartscore{} to measure and also to improve the faithfulness by designing an extractive-abstractive summarization model.

\paragraph{Extractive-abstractive (EA) summarization}
Various EA models have been developed for summarization  \citep{chen-bansal-2018-fast, mao2021dyle}. These methods perform unsupervised selection of sentences or phrases with the main goal of handling long documents. In this paper, however, we propose EA summarization as a means for improving faithfulness by leveraging human-annotated relevant sentences. Access to these sentences is crucial for improving faithfulness in summarization tasks with significant confounding and also greatly simplifies designing EA models. To the best of our knowledge, our work is the first that casts relevant sentence extraction as a seq2seq task.

\paragraph{Theoretical studies of summarization}
There have been limited theoretical studies of summarization. \citet{peyrard2019simple} quantify various aspects of summarization, namely redundancy, relevance, and informativeness, using information theoretic quantities.
\section{Conclusion}
In this paper, we explored methods for improving faithfulness and multiperspectivity of answer summarization models in community question answering. We theoretically and empirically showed that training models to generate summaries from a small number of human-annotated relevant sentences can significantly improve the faithfulness. Our theoretical result hold only when the dataset has the property that the final summary is independent of the original answers given a small number of relevant sentences. 


\section{Ethical Considerations}
\paragraph{Bias.} Biases may exists in the data that we used in the paper and models trained on these datasets can propagate these biases.

\paragraph{Environmental cost.} Models were trained on 8 GPUs on AWS nodes. Training baseline models took around a couple of hours while training the SURE-Summ model took 3-4 hours. In total we ran around 50 experiments including hyperparameter search for batch size, warmup steps, and threshold $\tau$ for UNSURE-Summ model. Steps were taken to minimize environmental cost by using small versions of pretrained models like t5-small and bart-base during model development. 

\paragraph{Model misuse.} Models trained in this paper can be expected to work as intended when evaluated on data similar to the one the models were trained on, e.g. other closely related CQA datasets. Our trained models cannot be expected to generalize to other domains or tasks. Even our best models, in spite of improving faithfulness over state of the art methods, still suffer from faithfulness issues and can hallucinate information. Further, research is needed to eliminate hallucination in abstractive summarization models. 
\bibliographystyle{acl_natbib}
\typeout{}
\bibliography{anthology,custom}
\clearpage
\begin{appendices}
\section{Theoretical result}
\label{app:theoretical_result}
\begin{proof}[Proof of Theorem \ref{thm:causal_effect}]
Since the causal diagram in Fig. \ref{fig:causal_model} is a Markov chain, 
from Section 6 of \citet{ay2008information} we have that $\MI(\Set{Q, X} \rightarrow Y) = MI(Q, X; Y)$
and $MI(\Set{Q, X_R} \rightarrow Y) = MI(\Set{Q, X_R} ; Y)$, where $MI$ denotes standard mutual information.
Therefore, we have that
\begin{align}
    &CE(X_{R^c}) \notag \\
    &\;= \Abs{MI(\Set{Q, X}; Y) - MI(\Set{Q, X_R} ; Y)} \label{eq:1a}
\end{align}
Next from chain rule of mutual information we have that,
\begin{align}
&MI(Y; \Set{X_R, X} \mid Q) \notag \\
    &\;= MI(Y; X_R \mid Q) + MI(Y; X \mid X_R, Q)  \label{eq:1b} \\
    &\;= MI(Y; X \mid Q) + MI(Y; X_R \mid X, Q) \label{eq:1c} 
\end{align}
Since $MI(Y; X \mid X_R, Q) = 0$, from \eqref{eq:1b}, \eqref{eq:1c}, and \eqref{eq:1a}
 we have that 
\begin{align*}
&H(X_R \mid X, Q) - H(X_R | X, Q, Y) \\
&\;= MI(Y; X_R \mid X, Q)  \\
&\;= MI(Y; X_R \mid Q) - MI(Y ; X \mid Q) \\
&\;= MI(Y; X_R \mid Q) + MI(Y; Q) \\
&\;\quad - MI(Y ; X \mid Q) - MI(Y; Q)\\
&\;= MI(Y; X_R, Q) - MI(Y ; X, Q) \\
&\;= CE(X_{R^c})
\end{align*}
where the last equality follows from chain rule of mutual information.
\end{proof}

\begin{proof}[Proof of Theorem \ref{thm:log_likelihood}]
\begin{align}
    l(f) &= - \Exp[(q,x,r,y) \sim \dP]{\log p(f(q, x) \mid x)} \\
    l(g) &= - \Exp[(q,x,r,y) \sim \dP]{\log p(g(q, x_r) \mid x_r)}
\end{align}
By the causal model in Fig. \ref{fig:causal_model} we have that $y = g(q,x_r) = f(q, x)$.
Therefore, 
\begin{align*}
&l(g) \\
&\;= -\Exp[(q,x,r,y) \sim \dP]{\log p(g(q, x_r) \mid q, x_r)} \\
&\;= -\Exp[(q,x,r,y) \sim \dP]{\log p(y \mid q, x_r)} \\
&\;= H(Y \mid Q, X_R)
\end{align*}
Similarly, $l(f) = H(Y \mid Q, X)$. Then 
\begin{align*}
&l(f) - l(g)  \\
&\;= H(Y \mid Q, X) - H(Y \mid Q, X_R) \\
&\;= H(Y \mid Q, X) - H(Y) \\
&\;\quad - H(Y \mid Q, X_R) + H(Y) \\
&\;= MI(Y; \Set{Q, X_R}) - MI(Y; \Set{Q, X}) \\
&\;= CE(X_{R^c}),
\end{align*}
where the last line follows from \eqref{eq:1a}.
\end{proof}

\section{Dataset details}
\label{app:dataset_details}
\begin{figure*}[t]
    \centering
    \begin{tcolorbox}[boxsep=0pt,left=0pt,colframe=white]
    \footnotesize
    \begin{tabular}{p{\linewidth}}
    \textbf{Question}: Every fall when the weather gets cooler it is a given that a mouse will end up in my house. I understand that this happens probably to most home owners ... But if I wanted to take some action next summer/early fall to prevent mice from entering, what could I do?\\
    \grayrule
    \textbf{Answers}:\\
    1. \textbf{This is probably a futile exercise}. Mice can squeeze through the smallest of gaps so you'd have to virtually hermetically seal your house to prevent any mouse coming in... \textbf{What you can do is make your house a less welcoming environment}. ...\\
    2. When I bought my house it had a big problem with mice. … What I did was: \textbf{make sure there was no accessible food} I checked every place in the house that a mice could hide behind. … \\
    3. \textbf{We live in the country and have a pretty significant mouse problem, we use a pest service and the first piece of advice the guy gave us was to take down (or move FAR away from the house) our bird feeder}. Birds eat the seed and spill about 2 seeds for every one they eat, mice LOVE bird feeders for a food source. ... \\
    4. We had a drop off in mice near our COOP building once a neighbor started feeding neighbor cats. There are a half dozen cats that show up in the early AM and stay near her home. Not sure if they are also hunting the mice or if the scent has caused the mice to seek other homes. \textbf{Sealing basement window cracks with silicone also helped}.\\
    5. \textbf{Keeping food away is probably the biggest thing you can do to keep mice (and other multi legged critters) away. You could probably put out poison, but not recommended if you're a pet owner, don't want any secondary poisoning. As a natural solution, you could attempt to attract hawks, falcons, owls or other birds of prey to the area}. I'm not an aviary expert but there's probably some way to attract them...\\
    6. \textbf{I've found no solutions to keeping them out, nor will I use poison, as I've no desire to see my dog or our neighbor's cat poisoned too. You can try to have your house sealed, but unless the seal is virtually hermetic, they will find their way into the warmth of your home}. And mice can chew holes in things, so maintaining the perfect seal will be tough. ...\\
    7. ... get some kitty pee : If you prefer, mice are supposed to hate the smell of peppermint and spearmint, so there are repellents made from mint :\\
    8. I was extremely skeptical of electronic pest repellers and I have no idea whether they work with rodents. However, much to my surprise, they work very well against roaches. ...  \\
    \grayrule
    \textbf{Cluster summaries}:\\
    1. It is widely considered that this is an almost impossible task.\\[2pt]
    2. You could make your home less attractive by making sure that any food is well sealed and out of reach.\\[2pt]
    3. Sealing cracks around windows could prevent them from entering the house.\\[2pt]
    4. It is possible to use mouse poison, but this could lead to poisoning other animals, like pets.\\[2pt]
    5. A rather extreme solution would be to attract birds of prey to the area, although it is not clear how this could be done.\\    
    \grayrule    
    \textbf{Summary}: Whilst you might attempt to seal your house as best as possible, the general consensus is that this is an almost impossible task. However, it is possible to make your home less attractive to mice by making sure that food is properly stored and out of reach. Mouse poison could be a potential solution, but there is the risk of poisoning other animals. A rather unusual solution could be to attract birds of prey to your garden, although, it's unclear how you could do this.\\
    \grayrule
    \textbf{BART}: The most effective way to stop mice from gaining access to your house is to block up any potential access holes or cover them with plastic. Failing this, you could also try putting out poison.
    \end{tabular}
    \end{tcolorbox} 
    \vspace{-0.5cm}
    \caption{The full example corresponding to Figure \ref{fig:example} from \answersumm{} dataset.}  
    \label{fig:full_example}
\end{figure*}
The \answersumm{} dataset contains 3131 examples in the training set, 500 examples in the validation set, and 1000 examples in the test set. 
Figure \ref{fig:full_example} shows the full example corresponding to the example shown in Figure \ref{fig:example}. A closer look at the BART generated summary reveals subtle intrinsic hallucination, where the model generated summary encourages the user to ``cover them [holes] with plastic'', whereas the original answers mention ``keep[ing] all food sealed in plastic containers'' and makes no mention of covering holes with ``plastic''. 
\section{Experiment Details}
\label{app:experiment_details}
Each model was trained using PyTorch distributed data parallel (DDP) training on 8 GPUs with batch size of 4 resulting in an effective batch size of 32. For models that ran out of memory with batch size 4 (repeated summarization models) we used batch size of 1 with 4 gradient accumulation steps to maintain effective batch size of 32. All models were trained for 50 epochs using Adam optimizer with learning rate of 1e-4 and a learning rate schedule with 1000 warmup steps and linear decay. We used the default generation parameters for the corresponding models to generate the final summaries. Model selection during training was done using ROUGE-1 score. We used the python rouge-score package for computing ROUGE score with stemming. Note that \citet{fabbri2021answersumm} use the files2rouge package (\url{https://github.com/pltrdy/files2rouge}) for computing ROUGE scores. They also compute ROUGE scores on the BPE tokenized reference and predicted summary which gives higher ROUGE-L scores.

\section{Details of human evaluation}
\label{app:human_eval}
Summaries are evaluated by annotators along the following five axes. Each axis is rated as 1 (worst) to 5 (best). We annotate each example by two people and average their rating.
\begin{itemize}
    \item Faithfulness: The summary should not manipulate the original answers' information (intrinsic hallucination) or invent novel facts (extrinsic hallucination). In other words, the summary should be entailed from the answers~\cite{maynez-etal-2020-faithfulness,kryscinski-etal-2020-evaluating}.
    \item Relevance: The summary should answer the query based on the provided answers~\cite{li-etal-2021-ease}.
    
    \item{Coherence}: Do the phrases of the summary make sense when put together?~\cite{grusky-etal-2018-newsroom}
    \item{Multiperspectivity}: Unlike the previous axes, this metric is specific to opinion summarization. The summary should include all the major  perspectives in the answers. Note that this differs with Relevance as summarizing only the top answer can be highly relevant but lacks the diversity and hence low multiperspectivity.
    \item{Overall Quality}: All things considered, how satisfying the answer summary  is?
\end{itemize}

Note that since we use pretrained language models, we did not observe grammatically to be an issue and hence, for the sake of brevity, removed this axis. Moreover, rare cases of this problem would be reflected in the overall quality.

Human annotation was performed by 34 professional raters employed by a US-based outsourcing company specializing in improving data for the development of ML and AI products and content moderation. All raters were native speakers of English: 25 have a BA (or higher) degree in Linguistics, 3 -- a BA in another related field, and 6 had a minimum education level of a high school diploma or equivalent. The raters compensation varied between \$8.7 and \$18 per hour (i.e., between 1.2 and 2.5 federal minimum wage) depending on the rater position and experience level. The data collection protocol was reviewed by an expert review board and secured expedited approval since the data presented to the raters contained no sensitive content. Participant consent was obtained as part of the non-disclosure agreement signed by each rater employee upon hire, which also included the statement that the collected materials may be used to assist in the development and advancement of software, machine learning models, artificial intelligence, algorithms and technology research. All raters also signed a sensitive content agreement that outlined the types of content they may be asked to work with during their employment, possible associated risks, and support and wellness resources provided by the company to its employees.

Prior to the task, the raters participated in a training session to become familiar with the evaluation instructions. These instructions included detailed definitions of evaluation criteria, contrasting examples for each criterion, and screenshots of the user interface to be used during data collection. During the course of data evaluation, the raters were presented with a user question, several answers/comments addressing this question, and the summary of those answers. The raters were further asked to read all the information provided to them carefully and rate each summary on a 5-point Likert scale on each of the 5 criteria discussed above (1 being the lowest degree of the given criterion and 5 being the highest). Evaluation of each summary across all 5 criteria took on average 8.5 minutes to complete, and each rater conducted an average of 216 evaluations over the course of 4 weeks of data collection. 

Each summary was evaluated by two individual raters. We assessed rater agreement for each of the 5 evaluation criteria by calculating percentages of cases where the assessments given by the two raters were either identical or within 1 point from one another. Out of 5 criteria, the raters demonstrated the highest level of agreement on Coherence (86.69\% of all assessments were in exact or close ($\pm$1) agreement). Relevance and Overall Quality both revealed the agreement level of 80.22\%. The agreement was lower for Multi-Perspectiveness (78.93\%) and Faithfulness (78.19\%). 

\end{appendices}

\end{document}